\documentclass[runningheads]{llncs}
\pdfoutput=1
\usepackage[utf8]{inputenc}
\usepackage{graphicx}
\usepackage{comment}
\usepackage{biblatex}
\usepackage{placeins}
\usepackage{amsmath,amssymb,amsfonts,bm} 
\usepackage{color}
\usepackage{booktabs}

\usepackage{import}
\usepackage{appendix}
\usepackage[caption=false]{subfig}
\usepackage{mathtools}
\usepackage{wrapfig}
\usepackage{hyperref}

\newcommand*\diff{\mathop{}\!\mathrm{d}}

\begin{document}

\pagestyle{headings}
\mainmatter
\def\ECCVSubNumber{5488}  

\title{Null-sampling for Interpretable \\and Fair Representations} 

\titlerunning{Null-sampling for Interpretable and Fair Representations}
%
\author{Thomas Kehrenberg \and
Myles Bartlett \and
Oliver Thomas \and \\
Novi Quadrianto}
\authorrunning{T. Kehrenberg et al.}
%
\institute{Predictive Analytics Lab (PAL), University of Sussex, Brighton, UK
\email{\{t.kehrenberg,m.bartlett,ot44,n.quadrianto\}@sussex.ac.uk}}
\maketitle

\begin{abstract}
\noindent
We propose to learn invariant representations, in the data domain, to achieve interpretability in algorithmic fairness. 
Invariance implies a selectivity for high level, relevant correlations w.r.t. class label annotations, and a robustness to irrelevant correlations with protected characteristics such as race or gender. 
We introduce a non-trivial setup in which the training set exhibits a strong bias such that class label annotations are irrelevant and spurious correlations cannot be distinguished.
To address this problem, we introduce an adversarially trained model with a \emph{null-sampling} procedure to produce invariant representations in the data domain.
To enable disentanglement, a partially-labelled \emph{representative} set is used.
By placing the representations into the data domain, the changes made by the model are easily examinable by human auditors.
We show the effectiveness of our method on both image and tabular datasets: Coloured MNIST, the CelebA and the Adult dataset.%
\footnote{The code can be found at \url{https://github.com/predictive-analytics-lab/nifr}.}
\end{abstract}

\begin{refsection}

\section{Introduction}

Without due consideration for the data collection process, machine learning algorithms can exacerbate biases, or even introduce new ones if proper control is not exerted over their learning \cite{holstein2019improving}. 
While most of these issues can be solved by
controlling and curating data collection in a fairness-conscious fashion, 
doing so is not always an option, such as when working with historical data.
Efforts to address this problem algorithmically have been centred on developing statistical definitions of fairness and learning models that satisfy these definitions.
One popular definition of fairness used to guide the training of fair classifiers, for example, is \emph{demographic parity}, stating that positive outcome rates should be equalised (or \emph{invariant}) across protected groups.

In the typical setup, we have an input $\bm{x}$, a sensitive attribute $s$ that represents some non-admissible information like gender
and a class label $y$ which is the prediction target.
The idea of fair \emph{representation} learning \cite{ZemWuSwePitetal13}\cite{edwardsstorkey}\cite{madras2018learning}
is then to transform the input $\bm{x}$ to a representation $\bm{z}$ which is invariant to $s$.
Thus, learning from $\bm{z}$ will not introduce a forbidden dependence on $s$.
A good fair representation is one that preserves most of the information from $\bm{x}$ while satisfying the aforementioned constraints.

As unlabelled data is much more freely available than labelled data, it is of interest to learn the representation in an unsupervised manner.
This will allow us to draw on a much more diverse pool of data to learn from.
While annotations for $y$ are often hard to come by (and often noisy \cite{KehCheQua18}),
annotations for the sensitive attribute $s$ are usually less so, as $s$ can often be obtained from demographic information provided by census data.
We thus consider the setting where the representation is learned from data
that is only labelled with $s$ and not $y$.
This is in contrast to most other representation learning methods.
We call the set used to learn the representation the \emph{representative} set,
because its distribution is meant to match the distribution of the deployment setting
(and is thus representative).

Once we have learnt the mapping from $\bm{x}$ to $\bm{z}$,
we can transform the \emph{training} set which, in contrast to the representative set, has the $y$ labels (and $s$ labels).
In order to make our method more widely applicable,
we consider an \emph{aggravated fairness problem}
in which the training set contains a strong spurious correlation between $s$ and $y$,
which makes it impossible to learn from it a representation which is invariant to $s$ but not invariant to $y$.
Non-invariance to $y$ is important in order to be able to predict $y$.
The training set thus does \emph{not} match the deployment setting,
thereby rendering the representative set essential for learning the right invariance.
From hereon, we will use the terms \emph{spurious} and \emph{sensitive} interchangeably, depending on the context, to refer to an attribute of the data we seek invariance to.
We can draw a connection between learning in the presence of spurious correlations and what \cite{kallus2018residual} call \emph{residual unfairness}.
Consider the Stop, Question and Frisk (SQF) dataset for example:
the data was collected in New York City, but the demographics of the recorded cases do not represent the true demographics of NYC well.
The demographic attributes of the recorded individuals might correlate so strongly with the prediction target that the two are nearly indistinguishable.
This is the scenario that we are investigating: $s$ and $y$ are so closely correlated in the labelled dataset that they cannot be distinguished, but the learning of $s$ is favoured due to being the ``path of least resistance''.
The deployment setting (i.e.\ the test set) does not possess this strong correlation and thus a na\"ive approach will lead to very unfair predictions.
In this case, a disentangled representation is insufficient;
the representation needs to be explicitly invariant solely with respect to $s$.
In our approach, we make use of the (partially labelled) representative set to learn this invariant representation.

While there is a substantial body of literature devoted to the problems of fair representation-learning, exactly how the invariance in question is achieved is often overlooked. 
When critical decisions, such as who should receive bail or be released from jail, are being deferred to an automated decision making system, it is critical that people be able to trust the logic of the model underlying it, whether it be via semantic or visual explanations. 
We build on the work of \cite{QuaShaTho19} and learn a decomposition ($f^{-1}: Z_s \times Z_{\neg s} \rightarrow X$) of the \emph{data domain} ($X$) into independent subspaces \emph{invariant} to  $s$ ($Z_{\neg s}$) and \emph{indicative} of $s$ ($Z_{s}$), which lends an interpretability that is absent from most representation-learning methods.
While model interpretability has no strict definition \cite{zhang2018visual}, we follow the intuition of \cite{adel2018discovering} -- \emph{a simple relationship to something we can understand}, a definition which representations in the data domain naturally fulfil.

Whether as a result of the aforementioned sampling bias or simply because the features necessarily co-occur, it is not rare for features to correlate with one another in real-world datasets.
Lipstick and gender for example, are two attributes that we expect to be highly correlated and to enforce invariance to gender can implicitly enforce invariance to makeup.
This is arguably the desired behaviour.
However, unforeseen biases in the data may engender cases which are less justifiable.
By baking interpretability into our model (by having representations in the data domain), though we still have no better control over what is learned, we can at least diagnose such pathologies.

To render our representations interpretable, we rely on a simple transformation we call \emph{null-sampling} to map invariant representations in the data domain.
Previous approaches to fair representation learning \cite{beutel,edwardsstorkey,madras2018learning,LouSweLi15} predominantly rely upon autoencoder models to jointly minimise reconstruction loss and invariance.
We discuss first how this can be done with such a model that we refer to as cVAE (conditional VAE),
before arguing that the bijectivity of invertible neural networks (INNs)~\cite{Dinh2014} makes them better suited to this task. 
We refer to the variant of our method based on these as cFlow (conditional Flow).
INNs have several properties that make them appealing for unsupervised representation learning. 
The focus of our approach is on creating invariant representations that preserve the non-sensitive information maximally, with only knowledge of $s$ and not of the target $y$, while at the same time having the ability to easily probe what has been learnt.



Our contribution is thus two-fold:
1) We propose a simple approach to generating representations 
that are invariant to a feature $s$, while having the benefit of interpretability that comes with being in the data domain.
We call our model \emph{NIFR} (\textbf{N}ull-sampling for \textbf{I}nterpretable and \textbf{F}air \textbf{R}epresentations).
2) We explore a setting where the labelled training set suffers from varying levels of sampling bias,
demonstrating an approach based on transferring information from a more diverse representative set,
with guarantees of the non-spurious information being preserved.

\begin{figure*}[tb]
  \centering
  \subfloat[Original images.]{%
      \scalebox{0.3}{\includegraphics[width=\textwidth]{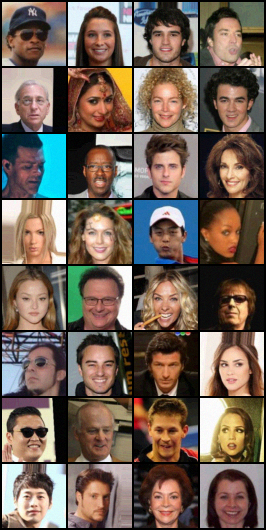}}%
      \label{fig:cflow_celeba_original_x}
  }
  \hfill
  \subfloat[$\bm{x}_u$ null-samples from the cFlow model.]{%
      \scalebox{0.3}{\includegraphics[width=\textwidth]{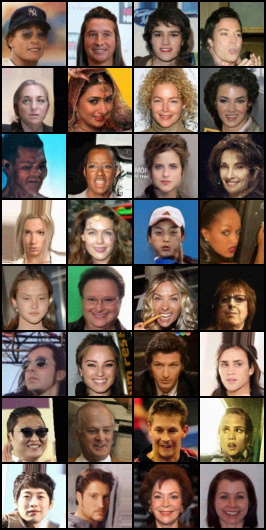}}%
      \label{fig:cflow_celeba_recon_y}
  }
  \hfill
  \subfloat[$\bm{x}_b$ null-samples from the cFlow model.]{%
      \scalebox{0.3}{\includegraphics[width=\textwidth]{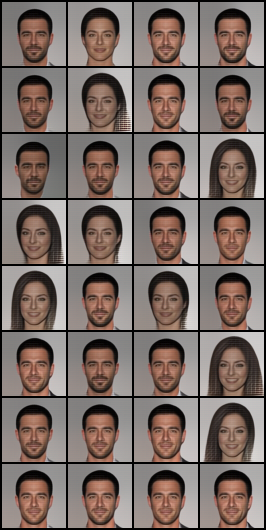}}%
      \label{fig:cflow_celeba_recon_s}
  }
  \caption{
    CelebA null-samples learned by our cFlow model, with gender as the sensitive attribute.
    (a) The original, untransformed samples from the CelebA dataset
    (b) Reconstructions using only information unrelated to $s$.
    (c) Reconstruction using only information related to $\neg s$.
    The model learns to disentangle gender from the non-gender related information.
    Note that some attributes like skin tone seem to change along with gender due to the correlation between the attributes.
    This is especially visible in images (1,1) and (3,2). Only because our representations are produced in the data-domain can we easily spot such instances of entanglement.
  }%
  \label{fig:celeba_cflow}
\end{figure*}

\begin{figure*}[!htb]
    \centering
    \subfloat[Samples from the cMNIST training set, $\sigma=0$.]{%
        \scalebox{0.3}{\includegraphics[width=\textwidth]{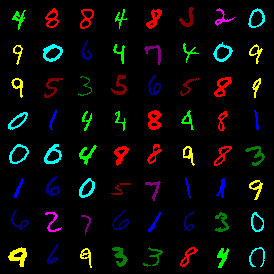}}%
        \label{fig:cflow_cmnist_task_train}
    }
    \hfill
    \subfloat[$x_u$ null-samples from the cFlow model.]{%
        \scalebox{0.3}{\includegraphics[width=\textwidth]{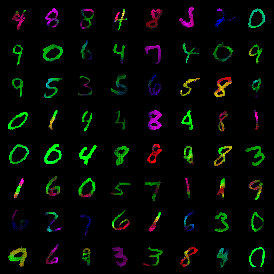}}%
        \label{fig:cflow_cmnist_y}
    }
    \hfill
    \subfloat[$x_b$ null-samples from the cFlow model.]{%
        \scalebox{0.3}{\includegraphics[width=\textwidth]{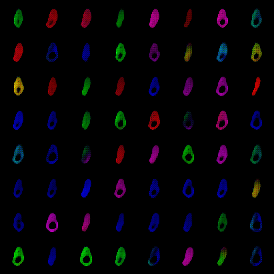}}%
        \label{fig:cflow_cmnist_s}
    }
    \caption{
        Sample images from the coloured MNIST dataset problem with $10$ predefined mean colours.
        (a): Images from the spuriously correlated subpopulation where colour is a reliable signal of the digit class-label.
        (b-c): Results of running our approach realised with cFlow on the cMNIST dataset.
        The model learns to retain the shape of the digit shape while removing the relationship with colour.
        A downstream classifier is now less prone to exploiting correlations between colour and the digit label class.
    }\label{fig:cmnist}
\end{figure*}

\section{Background}\label{sec:background}

\textbf{Learning fair representations.}
Given a sensitive attribute $s$ (for example, gender or race) and inputs $\bm{x}$,
a fair representation $\bm{z}$ of $\bm{x}$ is then one for which $\bm{z} \perp s$ holds,
while ideally also being predictive of the class label $y$.
\cite{ZemWuSwePitetal13} was the first to propose the learning of fair representations which allow for transfer to new classification tasks.
More recent methods are often based on variational autoencoders (VAEs)~\cite{kingma2013auto,LouSweLi15,edwardsstorkey,beutel}.
The achieved fairness of the representation can be measured with various fairness metrics.
These measure, however, usually how fair the predictions of a classifier are
and not how fair a representation is.

The appropriate measure of fairness for a given task is domain-specific \cite{liu2018delayed}
and there is often not a universally accepted measure.
However, \emph{Demographic Parity} is the most widely used~\cite{LouSweLi15,edwardsstorkey,beutel}.
Demographic Parity demands $\hat{y} \perp s$ where $\hat{y}$ refers to the predictions of the classifier.
In the context of fair representations, we measure the Demographic Parity of a downstream classifier, $f(\cdot )$, which is trained on the representation $z$ i.e.  $f: Z \to \hat{Y}$.

A core principle of all fairness methods is the \emph{accuracy-fairness trade-off}.
As previously stated, the fair representation should be invariant to $s$ ($\to$ fairness) but still be predictive of $y$ ($\to$ accuracy).
These desiderata cannot, in general, be simultaneously satisfied if $s$ and $y$ are correlated.

The majority of existing methods for fair representations also make use of $y$ labels during training,
in order to ensure that $\bm{z}$ remains predictive of $y$.
This aspect can, in theory, be removed from the methods,
but then there is no guarantee that information about $y$ is preserved \cite{LouSweLi15}. 

\textbf{Learning fair, transferrable representations.}
In addition to producing fair representations, \cite{madras2018learning} want to ensure the representations are transferrable.
Here, an adversary is used to remove sensitive information from a representation $z$.
Auxiliary prediction and reconstruction networks, to predict class label $y$
and reconstruct the input $x$ respectively,
are trained on top of $z$, with $s$ being ancillary input to the reconstruction.

Also related is \cite{creager2019flexibly} who employ a FactorVAE \cite{kim2018disentangling} regularised for fairness.
The idea is to learn a representation that is both disentangled and invariant to multiple sensitive attributes.
This factorisation makes the latent space easily manipulable such that the different subspaces can be freely removed and composed at test time.
Zeroing out the dimensions or replacing them with independent noise imparts invariance to the corresponding sensitive attribute.
This method closely resembles ours when we use an invertible encoder.
However, the emphasis of our approach is on interpretability, information-preservation, and coping with sampling bias - especially extreme cases where $|\, \textrm{supp}(S_{tr} \times Y_{tr}) | < |\, \textrm{supp}(S_{te} \times Y_{te}) |$.

Attempts were made by \cite{QuaShaTho19} prior to this work to learn fair representations in the data domain in order to make it interpretable and transferable.
In their work, the input is assumed to be additively decomposable in the feature space into a \emph{fair} and \emph{unfair} component, which together can be used by the decoder to recover the original input.
This allows us to examine representations in a human-interpretable space and confirm that the model is not learning a relationship reliant on a sensitive attribute.
Though a first step in this direction, we believe such a linear decomposition is not sufficiently expressive to fully capture the relationship between the sensitive and non-sensitive attributes.
Our approach allows for the modelling of more complex relationships.

\textbf{Learning in the presence of spurious correlations.}
Strong spurious correlations make the task of learning a robust classifier challenging: the classifier may learn to exploit correlations unrelated to the true causal relationship between the features and label, and thereby fail to generalise to novel settings.
This problem was recently tackled by \cite{ln2l} who apply a penalty based on the mutual information between the feature embedding and the spurious variable. 
While the method is effective under mild biasing, we show experimentally that it is not robust to the range of settings we consider.

Jacobsen et al. \cite{JacBehZemBet19} explore the vulnerability of traditional neural networks to spurious variables -- e.g., textures, in the case of ImageNet \cite{Geir18} -- and propose a INN-based solution akin to ours.
The INN's encoding is split such that one partition, $z_b$ is encouraged to be predictive of the spurious variable while the other serves as the logits for classification of the semantic label. 
Information related to the nuisance variable is ``pulled out'' of the logits as a result of maximising $\log p(s|z_n)$.
This specific approach, however, is incompatible with the settings we consider, due to its requirement that both $s$ and $y$ be available at training time.

Viewing the problem from a causal perspective, \cite{arjovsky2019invariant} develop a variant of empirical risk minimisation called invariant risk minimisation (IRM).
The goal of IRM is to train a predictor that generalises across a large set of unseen environments; because variables with spurious correlations do not represent a stable causal mechanism, the predictor learns to be invariant to them. IRM assumes that the training data is not \emph{iid} but is partitioned into distinct environments, $e \in E$. The optimal predictor is then defined as the minimiser of the sum of the empirical risk $R_e$ over this set. In contrast, we assume possession of only a single source of \emph{labelled}, albeit spuriously-correlated, data, but that we have a second source of data that is free of spurious correlations, with the benefit being that it only needs to be labelled \emph{with respect to $s$}.

\section{Interpretable Invariances by Null-Sampling}\label{ssec:general}
\begin{figure*}[tb]
    \centering
    \hfill
    \subfloat[cFlow model.]{%
        \scalebox{0.33}{\includegraphics[width=\textwidth]{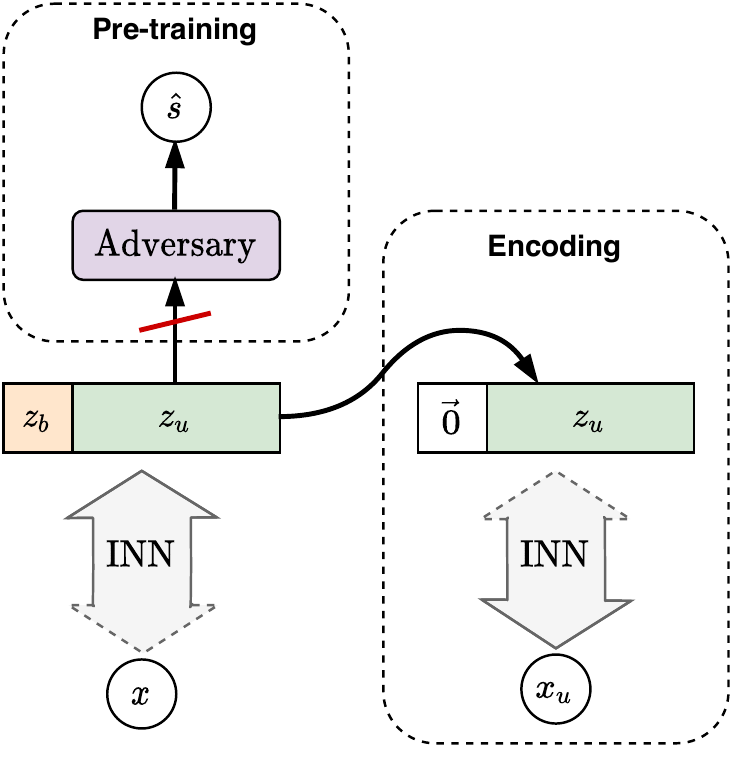}}%
        \label{fig:inn_diagram}
    }
    \hfill
    \subfloat[cVAE model.]{%
        \scalebox{0.4}{\includegraphics[width=\textwidth]{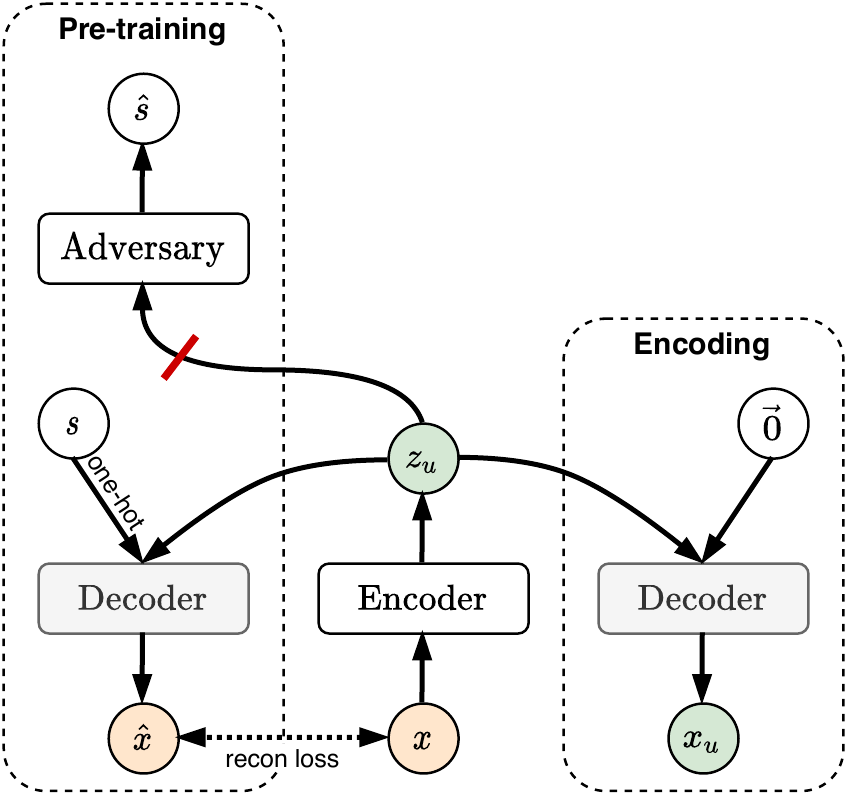}}%
        \label{fig:cvae_diagram}
    }
    \hfill
    \caption{
        Training procedure for our models. $x$: input, $s$: sensitive attribute, $z_u$: de-biased representation, $x_u$: de-biased version of the input in the data domain.
        The red bar indicates a gradient reversal layer, and $\stackrel{\rightarrow}{0}$ the null-sampling operation.
    }%
    \label{fig:model-diagrams}
\end{figure*}

\subsection{Problem Statement} 
\noindent We assume we are given inputs $\bm{x} \in \mathcal{X}$ and corresponding labels $y \in \mathcal{Y}$.
Furthermore, there is some spurious variable $s \in \mathcal{S}$ associated with each input $\bm{x}$ which we do \emph{not} want to predict. 
Let $X$, $S$ and $Y$ be random variables that take on the values $\bm{x}$, $s$ and $y$, respectively.
The fact that both $y$ and $s$ are predictive of $\bm{x}$ implies that $\mathcal{I}(X;Y), \mathcal{I}(X;S) > 0$, where $\mathcal{I}(\cdot ;\cdot)$ is the mutual information.
Note, however, that the conditional entropy is non-zero: $H(S|X) \neq 0$, i.e., $S$ is not completely determined by $X$.

The difficulty of this setup emerges in the training set: there is a close correspondence between $S$ and $Y$, such that for a model that sees the data through the lens of the loss function, the two are indistinguishable.
Furthermore, we assume that this is \emph{not} the case in the test set, meaning the model cannot rely on shortcuts provided by $S$ if it is to generalise from the training set.

We call this scenario where we only have access to the labels of a biasedly-sampled subpopulation
an \emph{aggravated fairness problem}.
These are not uncommon in the real-world.
For instance, in long-feedback systems such as mortgage-approval where the demographics of the subpopulation with observed outcomes is \emph{not} representative of the subpopulation on which the model has been deployed. 
In this case, $s$ has the potential to act as a false (or \emph{spurious}) indicator of the class label and
training a model with such a dataset would limit generalisability.
Let $(X^\mathit{tr}, S^\mathit{tr}, Y^\mathit{tr})$ then be the random variables sampled for the training set
and $(X^\mathit{te}, S^\mathit{te}, Y^\mathit{te})$ be the random variables for the test set.
The training and test sets thus induce the following inequality for their mutual information:
$\mathcal{I}(S^\mathit{tr}; Y^\mathit{tr}) \gg \mathcal{I}(S^\mathit{te}; Y^\mathit{te}) \approx 0$.

Our goal is to learn a representation $\bm{z}_u$ that is independent of $s$ and transferable between downstream tasks.
Complementary to $\bm{z}_u$, we refer to some abstract component of the model that absorbs the unwanted information related to $s$ as $\mathcal{B}$, the realisation of which we define with respect to each of the two models to be described.
The requirement for $\bm{z}_u$ can be expressed via mutual information:
\begin{align}
  I(\bm{z}_u;s) \overset{!}{=} 0~.
  \label{eq:migoal}
\end{align}
However, for the representation to be useful, we need to capture as much relevant information in the data as possible.
Thus, the combined objective function:
\begin{align}
  \min_{\theta} \mathbb{E}_{x \sim X}[-\log p_\theta(\bm{x})] + \lambda I(f_\theta(x);s)
  \label{eq:objectivetheory}
\end{align}
where $\theta$ refers to the trainable parameters of our model $f_\theta$ and $p_\theta(\bm{x})$ is the likelihood it assigns to the data.

We optimise this loss in an adversarial fashion by playing a min-max game, in which our encoder acts as the generative component.
The adversary is an auxiliary classifier $g$, which receives $\bm{z}_u$ as input and attempts to predict the spurious variable $s$.
We denote the parameters of the adversary as $\phi$;
for the parameters of the encoder we use $\theta$, as before.
The objective from Eq~\eqref{eq:objectivetheory} is then
\begin{align}
  \min_{\theta\in\Theta} \max_{\phi\in\Phi} \mathbb{E}_{x \sim X}[\log p_\theta(x) -\lambda\mathcal{L}_c(g_\phi(f_\theta(x))); s)]
  \label{eq:objectivepractical}
\end{align}
where $\mathcal{L}_c$ is the cross-entropy between the predictions for $s$ and the provided labels.
In practice, this adversarial term is realised with a gradient reversal layer (GRL) \cite{ganin2016domain} between $\bm{z}_u$ and $g$ as is common in adversarial approaches~\cite{edwardsstorkey}.

\subsection{The Disentanglement Dilemma} 
The objective in Eq~\eqref{eq:objectivepractical} balances the two desiderata: predicting $y$ and being invariant to $s$.
However, in the training set $(X^\mathit{tr}, S^\mathit{tr}, Y^\mathit{tr})$,
$y$ and $s$ are so strongly correlated that removing information about $s$ inevitably removes information about $y$.
This strong correlation makes existing methods fail under this setting.
In order to even define the right learning goal,
we require another source of information that allows us to disentangle $s$ and $y$.
For this, we assume the existence of another set of samples that follow a similar distribution to the test set,
but whilst the sensitive attribute is available, the class labels are not.
In reality, this is not an unreasonable assumption,
as, while properly annotated data is scarce, unlabelled data can be obtained in abundance (with demographic information from census data, electoral rolls, etc.).
Previous work has also considered treated ``unlabelled data'' as still having $s$ labels \cite{wick2019unlocking}.
We are restricted only in the sense that the spurious correlations we want to sever are indicated in the features.
We call this the \emph{representative set}, consisting of $X^\mathit{rep}$ and $S^\mathit{rep}$.
It fulfils $\mathcal{I}(S^\mathit{rep}; Y^\mathit{rep}) \approx 0$
(or rather, it would, if the class labels $Y^\mathit{rep}$ were available).

We now summarise the training procedure; an outline for the invertible network model (cFlow) can be seen in Fig.~\ref{fig:inn_diagram}.
First, the encoder network $f$ is trained on ($X^\mathit{rep}, S^\mathit{rep}$), during the first phase.
The trained network is then used to encode the training set,
taking in $\bm{x}$ and producing the representation, $\bm{z}_u$, decorrelated from the spurious variable.
The encoded dataset can then be used to train any off-the-shelf classifier safely, with information about the spurious variable having been absorbed by some auxiliary component $\mathcal{B}$.
In the case of the conditional VAE (cVAE) model,
$\mathcal{B}$ takes the form of the decoder subnetwork, which reconstructs the data conditional on a one-hot encoding of $s$,
while for the invertible network $\mathcal{B}$ is realised as a partition of the feature map $\bm{z}$
(such that $\bm{z} = [\bm{z}_u, \bm{z}_b]$), given the bijective constraint.
Thus, the classifier cannot take the shortcut of learning $s$ and instead must learn how to predict $y$ directly.
Obtaining the $s$-invariant representations, $\bm{x}_u$, in the data domain
is simply a matter of replacing the $\mathcal{B}$ component of the decoder's input for the cVAE,
and $\bm{z}_b$ for cFlow, with a zero vector of equivalent size.
We refer to this procedure used to generate $\bm{x}_u$ as \emph{null-sampling} (here, with respect to $\bm{z}_b)$.

Null-sampling resembles the \emph{annihilation} operation described in \cite{xiao2017dna}, however we note that the two serve very different roles.  Whereas the annihilation operation serves as a regulariser to prevent trivial solutions (similar to \cite{jaiswal2018unsupervised}), null-sampling is used to generate the invariant representations post-training.

\subsection{Conditional Decoding}%
\label{conddec}
\noindent We first describe a VAE-based model similar to that proposed in~\cite{madras2018learning}, before highlighting some of its shortcomings that motivate the choice of an invertible representation learner.

The model takes the form of a class conditional $\beta$-VAE \cite{higgins2017beta}, in which the decoder is conditioned on the spurious attribute.
We use $\theta_{enc}, \theta_{dec} \in \theta$ to denote the parameters of the encoder and decoder sub-networks, respectively.
Concretely, the encoder component performs the mapping $x \rightarrow{\bm{z}_u}$, while $\mathcal{B}$ is instantiated as the decoder,
$\mathcal{B} \coloneqq p_{\theta_{dec}}(x|z_u, s)$, which takes in a concatenation of the learned non-spurious latent vector $\bm{z}_u$ and a one-hot encoding of the spurious label $s$ to produce a reconstruction of the input $\hat{x}$.
Conditioning on a one-hot encoding of $s$, rather than a single value, as done in \cite{madras2018learning} is the key to visualising invariant representations in the data domain.
If $\mathcal{I}(z_u; s)$ is properly minimised, the decoder can only derive its information about $s$ from the label, thereby freeing up $\bm{z}_u$ from encoding the unwanted information while still allowing for reconstruction of the input.
Thus, by feeding a zero-vector to the decoder we achieve $\hat{x} \perp s$.
The full learning objective for the cVAE is given as
\begin{align}
\begin{split}
    \mathcal{L}_{\mathrm{cVAE}} =& \mathbb{E}_{q_{\theta_{enc}}(z_u, b|x)}[\log p_{\theta_{dec}}(x|z, b) - \log p_{\theta_{dec}}(s|z_u)] \\ 
    &- \beta D_{KL}(q_{\theta_{enc}}(z_u |x) \| p(z_u))
\end{split}
\end{align}
where $\beta$ is a hyperparameter that determines the trade-off between reconstruction accuracy and independence constraints,
and $p(\bm{z}_u)$ is the prior imposed on the variational posterior.
For all our experiments, $p(\bm{z}_u)$ is realised as an Isotropic Gaussian.
Fig.~\ref{fig:cvae_diagram} summarises the procedure as a diagram.

While we show this setup can indeed work for simple problems, as~\cite{madras2018learning} before us have, we show that it lacks scalability due to disagreement between the components of the loss.
Since information about $s$ is only available to the decoder as a binary encoding,
if the relationship between $s$ and $x$ is highly non-linear and cannot be summarised by a simple on/off mechanism, as is the case if $s$ is an attribute such as gender,
off-loading information to the decoder by conditioning is no longer possible. As a result, $\bm{z}_u$ is forced to carry information about $s$ in order to minimise the reconstruction error. 

The obvious solution to this is to allow the encoder to store information about $s$ in a partition of the latent space as in  \cite{creager2019flexibly}.  However, we question whether an autoencoder is the best choice for this setup, with the view that an invertible model is the better tool for the task. Using an invertible model has several guarantees, namely complete information-preservation and freedom from a reconstruction loss, the importance of which we elaborate on below.

\subsection{Conditional Flow}\label{cflow}
\textbf{Invertible Neural Networks.}
Invertible neural networks are a class of neural network architecture characterised by a bijective mapping between their inputs and output \cite{Dinh2014}. The transformations are designed such that their inverses and Jacobians are efficiently computable.
These flow-based models permit \emph{exact} likelihood estimation \cite{normflows2015} through the warping of a base density with a series of invertible transformations and computing the resulting, highly multi-modal, but still normalised, density, using the change of variable theorem:
\begin{align}
\begin{split}
  \log p(x) &= \log p(z) + 
   \sum \log \left| \det\left( \frac{\diff h_i}{ h_{i-1}}\right) \right|, 
  \quad p(z) = \mathcal{N}(z; 0, \mathbb{I})
  \label{eq:changeofvariables}
\end{split}
\end{align}
where $h_i$ refers to the outputs of the layers of the network and $p(z)$ is the base density, specifically an Isotropic Gaussian in our case.
Training of the invertible neural network is then reduced to maximising $\log p(x)$ over the training set,
i.e.\ maximising the probability the network assigns to samples in the training set.

\textbf{The Benefits of Bijectivity.}
Using an invertible network to generate our encoding, $\bm{z}_u$, carries a number of advantages over other approaches.
Ordinarily, the main benefit of flow-based models is that they permit exact density estimation. 
However, since we are not interested in sampling from the model's distribution, in our case the likelihood term serves as a regulariser, as it does for  \cite{JacSmeOya18}. 
Critically, this forces the mean of each latent dimension to zero enabling null-sampling. 
The invertible property of the network guarantees the preservation of all information relevant to $y$ which is independent of $s$, regardless of how it is allocated in the output space.
Secondly, we conjecture that the encodings are more robust to out-of-distribution data.
Whereas an autoencoder could map a previously seen input and a previously unseen input to the same representation,
an invertible network sidesteps this due to the network's bijective property, ensuring all relevant information is stored somewhere. This opens up the possibility of transfer learning between datasets with a similar manifestation of $s$, as we demonstrate in the Appendix~\ref{sec:transfer-learning}.

Under our framework, the invertible network $f$ maps the inputs $\bm{x}$ to a representation $\bm{z}_u$:
$f(\bm{x}) = \bm{z}$.
We interpret the embedding $\bm{z}$ as being the concatenation of two smaller embeddings: $\bm{z} = [\bm{z}_u, \bm{z}_b]$.
The dimensionality of $\bm{z}_b$, and $\bm{z}_u$, by complement, is a free parameter (see Appendix~\ref{ssec:partition-size-matters} for tuning strategies).
As $f$ is invertible, $\bm{x}$ can be recovered like so:
\begin{align}
  \bm{x} = f^{-1}([\bm{z}_u, \bm{z}_b])
  \label{eq:zreconstruct}
\end{align}
where $\bm{z}_b$ is required for equality of the output dimension and input dimension to satisfy the bijectivity of the network -- we cannot output $\bm{z}_u$ alone, but have to output $\bm{z}_b$ as well. In order to generate the pre-image of $\bm{z}_u$, we perform null-sampling with respect to $\bm{z}_b$ by zeroing-out the elements of $\bm{z}_b$ (such that $\bm{x}_{u} = f^{-1}([\bm{z}_{u}, \stackrel{\rightarrow}{0}])$), i.e. setting them to the mean of the prior density, $\mathcal{N}(z;0, I)$.

How can we be sure that $\bm{z}_u$ contains enough information about $y$?
The importance of the invertible architecture bears out from this consideration. 
As long as $\bm{z}_b$ does not contain the information about $y$, $\bm{z}_u$ necessarily must.
We can raise or lower the information capacity of $\bm{z}_b$ by adjusting its size;
this should be set to the smallest size sufficient to capture all information about $s$, so as not to sacrifice class-relevant information.
Appendix~\ref{ssec:size-of-zb} explores the effects of the size further.






\section{Experiments}
\noindent We present experiments to demonstrate that the null-sampled representations are in fact invariant to $s$
while still allowing a classifier to predict $y$ from them.
We run our cVAE and cFlow models on the coloured MNIST (cMNIST) and CelebA dataset,
which we artificially bias, first describing the sampling procedure we follow to do so for non-synthetic datasets.
As baselines we have the model of~\cite{ln2l} (Ln2L) and the same CNN used to evaluate the cFlow and cVAE models
but with the unmodified images as input (CNN).
For the cFlow model we adopt a Glow-like architecture~\cite{KinDha18},
while both subnetworks of the cVAE model comprise gated convolutions~\cite{van2016conditional}, where the encoding size is $256$.
For cMNIST, we construct the Ln2L baseline according to its original description, for CelebA,
we treat it as an augmentation of the baseline CNN's objective function.
Detailed information regarding model architectures can be found in Appendix~\ref{sec:architectures} and~\ref{sec:optimisation-details}.
\begin{figure}[tb]
    \centering
    \includegraphics[width=0.7\textwidth]{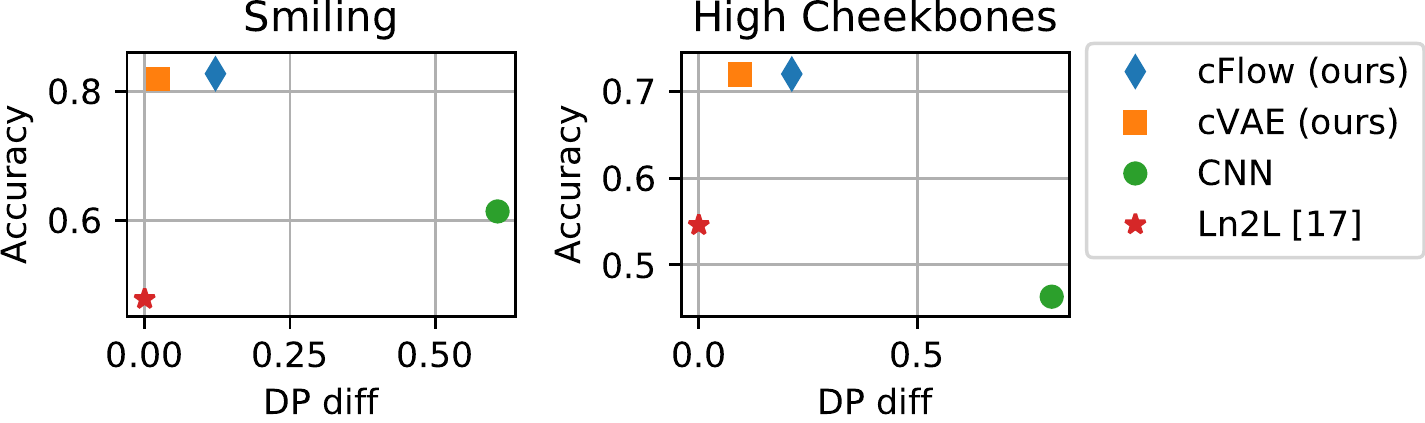}
    \caption{
        Performance of our model for different targets (mixing factor $\eta=0$).
        Left: \emph{Smiling} as target, right: \emph{high cheekbones}.
        \emph{DP diff} measures fairness with respect to demographic parity.
        A perfectly fair model has a \emph{DP diff} of 0.
    }%
    \label{fig:celeba-targets}
\end{figure}

\begin{figure}[tb]
    \centering
    \includegraphics[width=0.7\textwidth]{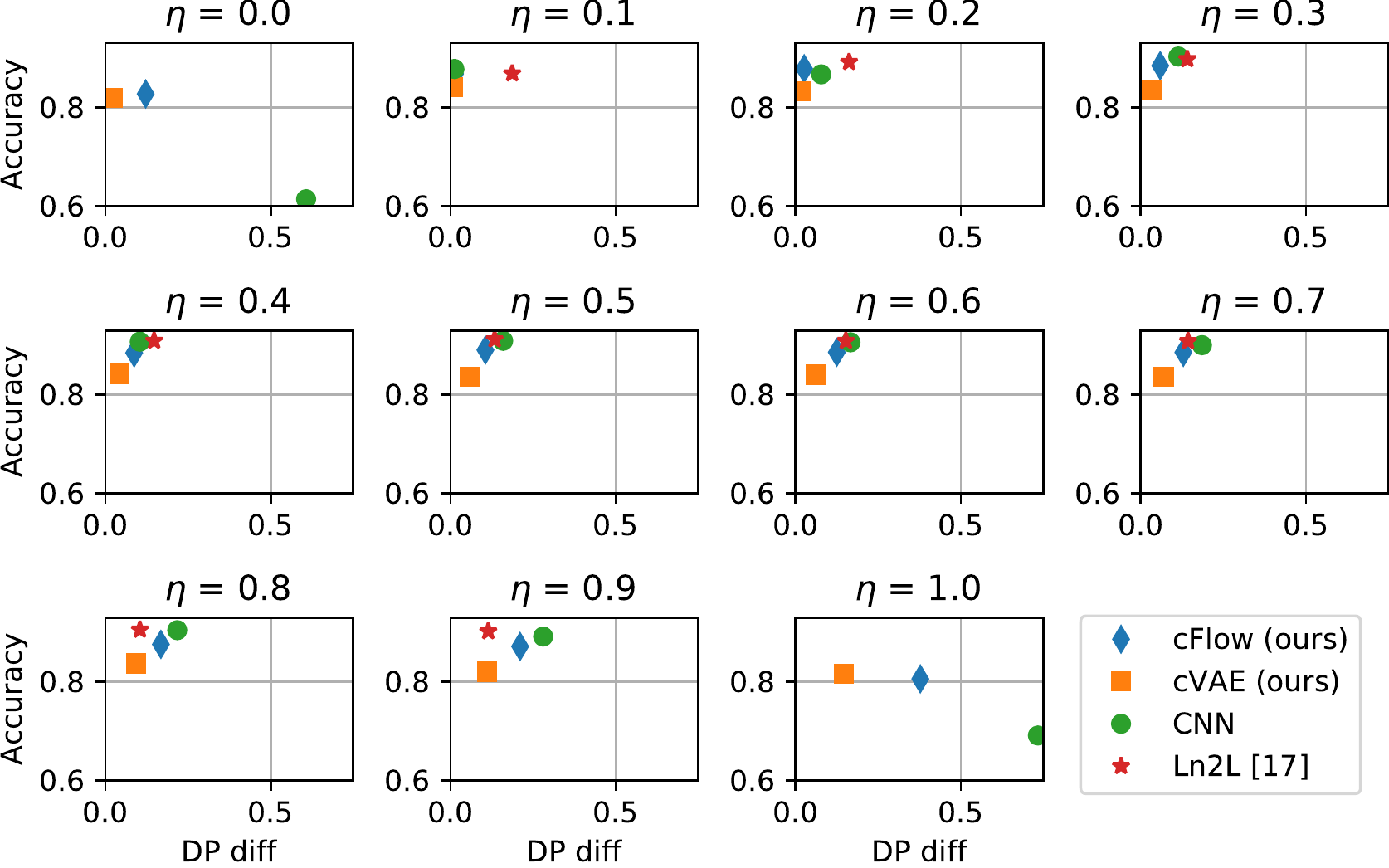}
    \caption{
        Performance of our model for the target ``smiling'' for different mixing factors $\eta$.
        \emph{DP diff} measures fairness with respect to demographic parity.
        A perfectly fair model has a \emph{DP diff} of 0, thus the closer to top-left the better it is in terms of we accuracy-fairness trade-off.
        Only values $\eta=0$ and $\eta=1$ correspond to the scenario of a strongly biased training set.
        The results for $0.1\leq \eta\leq 0.9$ are to confirm that our model does not harm performance for non-biased training sets.
    }%
    \label{fig:celeba-multiplot}
\end{figure}

\textbf{Synthesising Dataset Bias.}
\noindent For our experiments, we require a training set that exhibits a strong spurious correlation, together with a test set that does not.
For cMNIST, this is easily satisfied as we have complete control over the data generation process.
For CelebA and  UCI Adult, on the other hand,
we have to generate the split from the existing data.
To this end, we first set aside a randomly selected portion of the dataset from which to sample the biased dataset
The portion itself is then split further into two parts:
one in which $(s=-1 \land y=-1) \lor (s=+1 \land y=+1)$ holds true for all samples, call this part $\mathcal{D}_{eq}$,
and the other part, call it $\mathcal{D}_{opp}$, which contains the remaining samples.
To investigate the behaviour at different levels of correlation,
we mix these two subsets according to a mixing factor $\eta$.
For $\eta \leq \tfrac{1}{2}$, we combine (all of) $\mathcal{D}_{eq}$
with a fraction of $2\eta$ from $\mathcal{D}_{opp}$.
For $\eta > \tfrac{1}{2}$, we combine (all of) $\mathcal{D}_{opp}$
and a fraction of $2(1 -\eta)$ from $\mathcal{D}_{eq}$.
Thus, for $\eta=0$, the biased dataset is just $\mathcal{D}_{eq}$,
for $\eta=1$ it is just $\mathcal{D}_{opp}$
and for $\eta=\tfrac{1}{2}$ the biased dataset is an ordinary subset of the whole data. The test set is simply the data remaining from the initial split.

\textbf{Evaluation protocol.}
We evaluate our results in terms of accuracy and fairness.
A model that perfectly decouples its predictions from $s$ will achieve near-uniform accuracy across all biasing-levels.
For binary $s$/$y$ we quantify the fairness of a classifier's predictions using \emph{demographic parity} (DP): the  absolute difference in the probability  of a positive prediction for each sensitive group.


\begin{figure}[tb]
    \centering
    \includegraphics[width=0.7\textwidth]{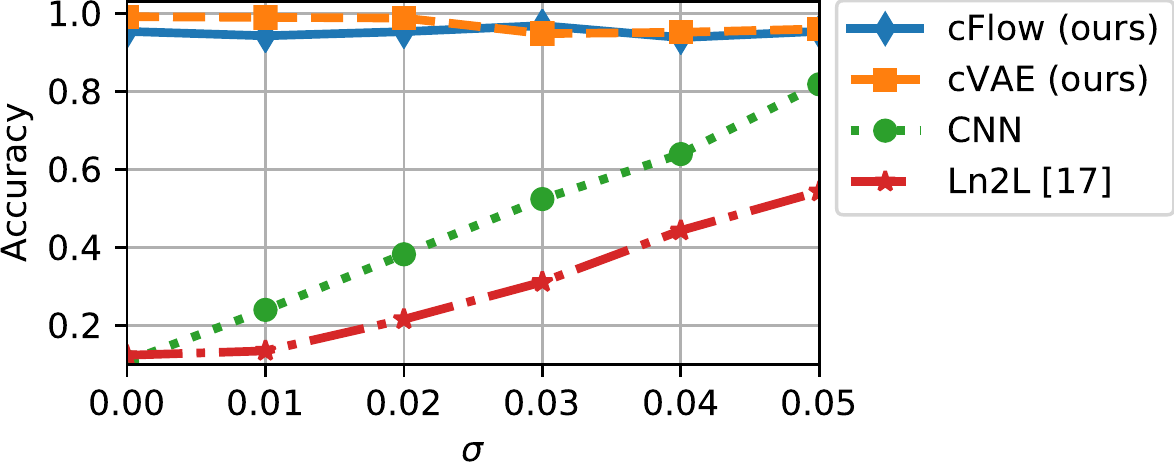}
    \caption{
        Accuracy of our approach in comparison with other baseline models on the cMNIST dataset, for different standard deviations ($\sigma$) for the colour sampling.
    }%
    \label{fig:cmnist_chart}
\end{figure}
\subsection{Experimental results}
We report the results from two image datasets.
cMNIST, a synthetic dataset, is a good starting point for evaluating our model due to the direct control we have over the biasing.
CelebA, on the other hand, is a more practical and challenging example.
We also test our method on a tabular dataset, the Adult dataset.

\textbf{cMNIST.}\label{ssec:cmnist}
The coloured MNIST (cMNIST) dataset is a variant of the MNIST dataset in which the digits are coloured.
In the training set, the colours have a one-to-one correspondence with the digit class.
In the test set (and the representative set), colours are assigned randomly.
The colours are drawn from Gaussians with 10 different means.
We follow the colourisation procedure outlined by~\cite{ln2l}, with the mean colour values selected so as to be maximally dispersed.
The full list of such values can be found in Appendix~\ref{sec:color-details}.
We produce multiple variants of the cMNIST dataset corresponding to different standard deviations $\sigma$ for the colour sampling:
$\sigma \in \{0.00, 0.01, ..., 0.05 \}$.

For this specific dataset, we can establish an additional baseline by simply grey-scaling the dataset
which only leaves the luminosity as spurious information.
We also evaluate the model, with all the associated hyperparameters, from~\cite{ln2l}.
The only difference between the setups is the dataset creation, including the range of $\sigma$ values we consider.
Our versions of the dataset, on the whole, exhibit much stronger colour bias, to the point of the mapping the digit's colour and class being bijective.
Fig.~\ref{fig:cmnist_chart} shows that the model significantly underperforms even the na\"ive baseline, aside from at $\sigma = 0$, where they are on par.

Inspection of the null-samples shows that both the cVAE and cFlow model succeed in removing almost all colour information, which is supported quantitatively by Fig.~\ref{fig:cmnist_chart}.
While the cVAE outperforms cFlow marginally at low $\sigma$ values, performance degrades
as this increases.
This highlights the problems with the conditional decoder we anticipated in Section~\ref{conddec}.
The lower $\sigma$, and therefore the variation in sampled colour, is, the more reliably the $s$ label, corresponding to the mean of RGB distribution, encodes information about the colour.
For higher $\sigma$ values, the sampled colours can deviate far from the mean and so the encoder must incorporate information about $s$ into its representation if it is to minimise the reconstruction loss.
cFlow, on the other hand, is consistent across $\sigma$ values.

\textbf{CelebA.}
To evaluate the effectiveness of our framework on real-world image data we use the CelebA dataset~\cite{liu2015faceattributes}, consisting of 202,599 celebrity images.
These images are annotated with various binary physical  attributes, including ``gender'', ``hair color'', ``young'', etc, from which we  select our sensitive and target attributes.
The images are centre cropped and resized to $64\times64$, as is standard practice.
For our experiments, we designate ``gender'' as the sensitive attribute,
and ``smiling'' and ``high cheekbones'' as target attributes.
We chose gender as the sensitive attribute as it a common sensitive attribute in the fairness literature.
For the target attributes, we chose attributes that are harder to learn than gender and which do not correlate too strongly with gender in the dataset
(``wearing lipstick'' for example being an attribute too closely correlated with gender).
The model is trained on the representative set (normal subset of CelebA)
and is then used to encode the artificially biased training set and the test set.
The results for the most strongly biased training set ($\eta=0$) can be found in Fig.~\ref{fig:celeba-targets}.
Our method outperforms the baselines in accuracy and fairness.

We also assess performance for different mixing factors ($\eta$) which correspond to varying degrees of bias in the training set
(see Fig.~\ref{fig:celeba-multiplot}).
This is to verify that the model does not \emph{harm} performance when there is not much bias in the training set.
For these experiments, the model is trained once on the representative set and is then used to encode different training sets.
The results show that for the intermediate values of $\eta$, our model incurs a small penalty in terms of accuracy,
but at the same time makes the results \emph{fairer} (corresponding to an accuracy-fairness trade-off). Qualitative results can be found in Fig.~\ref{fig:celeba_cflow} (images from cVAE can be found in Appendix~\ref{sec:qual-results-celeba}).

To show that our method can handle multinomial, as well as binary, sensitive attributes, we also conduct experiments with $s=\textrm{hair color}$ as a ternary attribute (``Blonde'' ``Black'', ``Brown''), excluding ``Red'' because of the paucity of samples and the noisiness of their labels. The results for these experiments can be found in Appendix~\ref{ssec:multi-s}.

\begin{wrapfigure}{l}{0.6\linewidth}
\centering
  \includegraphics[width=\linewidth]{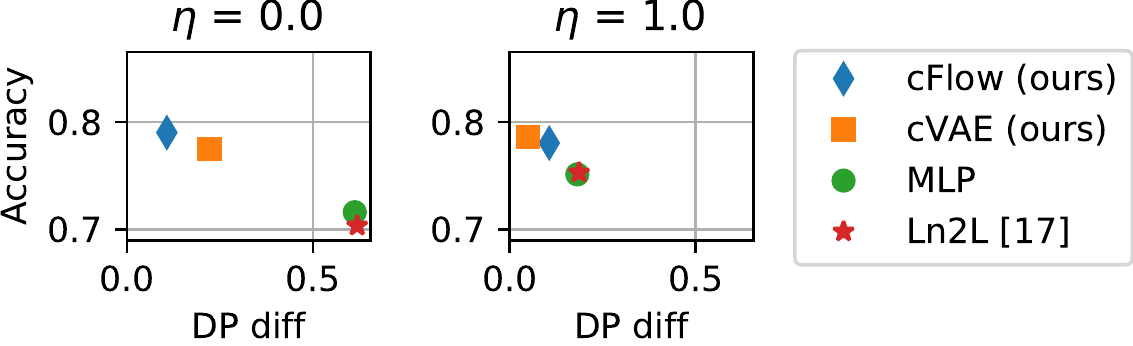}
  \caption{
      Results for the \textbf{Adult} dataset.
      The $x$-axis corresponds to the difference in positive rates.
      An ideal result would occupy the \textbf{top-left}.
  }%
  \label{fig:adult-chart}
\end{wrapfigure}

\noindent\textbf{Results for the UCI Adult dataset.}
The UCI Adult dataset consists of census data and is commonly used to evaluate models focused on algorithmic fairness.
Following convention, we designate ``gender'' as the sensitive attribute $s$ and whether an individual's salary is \$50,000 or greater as $y$.
We show the performance of our approach in comparison to baseline approaches in Fig. \ref{fig:adult-chart}.
We evaluate the performance of all models for mixing factors ($\eta$) $0$ and $1$. 
Results shown in Fig. \ref{fig:adult-chart} show that
we match or exceed the baseline.
In terms of fairness metrics, our approach generally outperforms the baseline models for both of $\eta$.
Detailed results can be found in the Appendix~\ref{ssec:detailed-adult}.

We also did experiments to show that the encoder transfers to other tasks. These transfer-learning experiments can be found in Appendix~\ref{sec:transfer-learning}.


\section{Conclusion}\label{sec:conclusion}
We have proposed a general and straightforward framework for producing invariant representations, under the assumption that a representative but partially-labelled \emph{representative} set is available.
Training consists of two stages:
an encoder is first trained on the representative set to produce a representation that is invariant to a designated spurious feature. 
This is then used as input for a downstream task-classifier, the training data for which might exhibit extreme bias with respect to that feature.
We train both a VAE- and INN-based model according to this procedure, and show that the latter is particularly well-suited to this setting due to its losslessness. 
The design of the models allows for representations that are in the data domain and therefore exhibit meaningful invariances. 
We characterise this for synthetic as well as real-world datasets for which we develop a method for simulating sampling bias.

\section*{Acknowledgements}
This work was in part funded by the European Research Council 
under the ERC grant agreement no. 851538.
We are grateful to NVIDIA for donating GPUs.
\FloatBarrier%
\printbibliography%
\end{refsection}
\clearpage

\pagestyle{headings}
\authorrunning{T. Kehrenberg et al.}
\titlerunning{Null-sampling for Interpretable and Fair Representations}
\backmatter%

\begin{refsection}

\title{Appendix}
\author{}
\institute{}
\maketitle
\thispagestyle{headings}


\begin{appendix}

\section{Model Architectures}%
\label{sec:architectures}
\noindent For both cMNIST and CelebA we parameterise the coupling layers with the same convolutional architecture as in \cite{KinDha18}, consisting of $3$ convolutional layers each with $512$ filters of, in order, sizes $3\times3$, $1\times1$, and $3\times3$.
Following \cite{ardizzone2019guided}, we Xavier initialise all but the last convolutional layer of the $s$ and $t$ sub-networks which itself is zero-initialised so that the coupling layers begin by performing an identity transform. We used a Glow-like architecture \cite{KinDha18} (affine coupling layers together with checkerboard reshaping and invertible $1\times1$ convolutions) for the convolutional INNs. Table \ref{tab:inn_architectures} summarises the INN architectures used for each dataset.

For the image datasets each level of the cVAE encoder consists of two gated convolutional layers \cite{van2016conditional} with ReLU activation. 
At each subsequent level, the number of filters is doubled, starting with an initial value 32 and 64 in the case of CelebA and cMNIST respectively. 
In the case of the Adult dataset, we use an encoder with one fully-connected hidden layer of width $35$, followed by SeLU activation \cite{klambauer2017self}. 
For both cMNIST and CelebA, we downsample to a feature map with spatial dimensions $8\times8$, but with $3$ and $16$ channels respectively. 
For the Adult dataset, the encoding is a vector of size $35$. 
The output layer specifies both the parameters (mean and variance) of the representation's distribution. 
In all cases the KL-divergence is computed with respect to a standard isotropic Gaussian prior. 
Details of the encoder architectures can be found in Table \ref{tab:vae_architectures}. The loss pre-factors were sampled from a logarithmic scale; without proper balancing the networks can exhibit instability, especially during the early stages of training.

\begin{table*}[ht]
\caption{INN architecture used for each dataset.}
\label{tab:inn_architectures}
\centering
\begin{tabular*}{\textwidth}{l@{\extracolsep{\fill}}llllll}
\toprule
\multicolumn{1}{l}{Dataset} & Levels & Level depth & Coupling channels & Input to discriminator(s) \\ \midrule
UCI Adult                   & 1      & 1     & 35       & Null-samples       \\
cMNIST                      & 3      & 16     & 512      & Encodings               \\
CelebA                      & 3      & 32     & 512      & Encodings        \\ \bottomrule
\end{tabular*}
\end{table*}

\begin{table}[ht]
\caption{cVAE encoder architecture used for each dataset. The decoder architecture in each case mirrors that of its encoder counterpart through use of transposed convolutions. For the adult dataset we apply $\ell_2$ and cross-entropy losses to the reconstructions of the continuous features and discrete features, respectively.}
\label{tab:vae_architectures}
\centering
\begin{tabular*}{\textwidth}{l@{\extracolsep{\fill}}lllll}
\toprule
Dataset   & Initial channels & Levels & $\beta$ & Recon. loss \\
\midrule
UCI Adult & 35               & --     & 0       & $\ell_2$ + CE\\
cMNIST    & 32               & 4      & 0.01    & $\ell_2$ \\
CelebA    & 32               & 5      & 1       & $\ell_1$ \\ 
\bottomrule
\end{tabular*}
\end{table}

\section{Additional results}\label{sec:additional-results}
\subsection{Detailed results for UCI Adult dataset}\label{ssec:detailed-adult}
\begin{figure}[htb]
  \centering
  \includegraphics[width=0.85\textwidth]{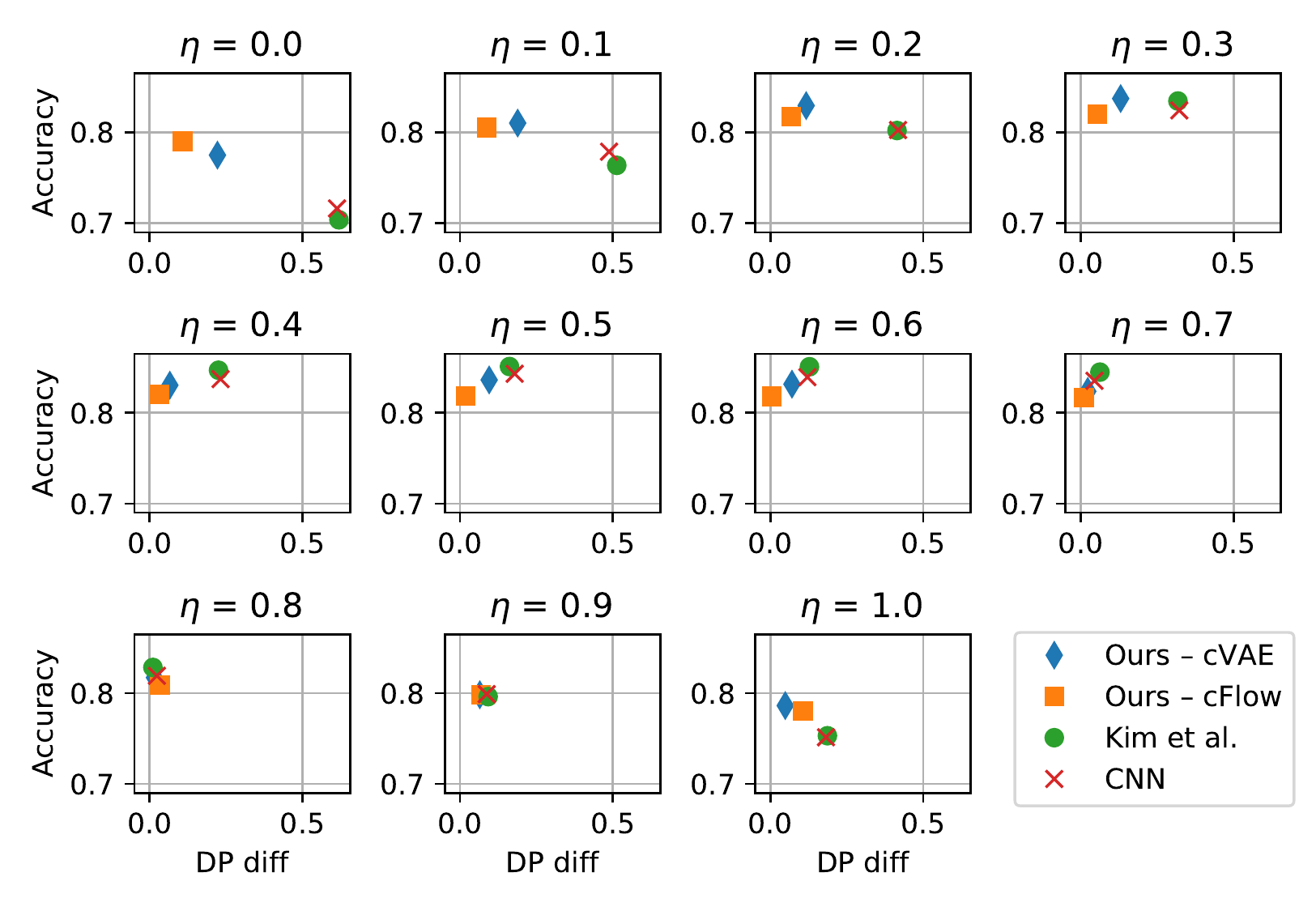}
  \caption{
      Results for the \textbf{Adult} dataset.
      The $x$-axis corresponds to the difference in positive rates.
      An ideal result would occupy the \textbf{top-left}.
  }%
  \label{fig:big-adult-chart}
\end{figure}

This census data is commonly used to evaluate models focused on algorithmic fairness.
Following convention, we designate ``gender'' as the $s$ and whether an individual's salary is \$50,000 or greater as $y$.
We show the performance of our approach in comparison to baseline approaches in Fig.~\ref{fig:big-adult-chart}.
We evaluate the performance of all models for mixing factors ($\eta$) of value $\{0, 0.1, ..., 1\}$. 
Results shown in Fig.~\ref{fig:big-adult-chart} show that whilst our model fails to surpass the baseline models in terms of accuracy for the balanced case (and those close to it), we match or exceed the baseline as $\eta $ moves the dataset to a more imbalanced setting. In terms of fairness metrics,  our approach generally outperforms the baseline models regardless of $\eta$.

\subsection{Multinomial sensitive attributes}\label{ssec:multi-s}
\begin{figure}[htb]
    \centering
    \includegraphics[width=0.7\textwidth]{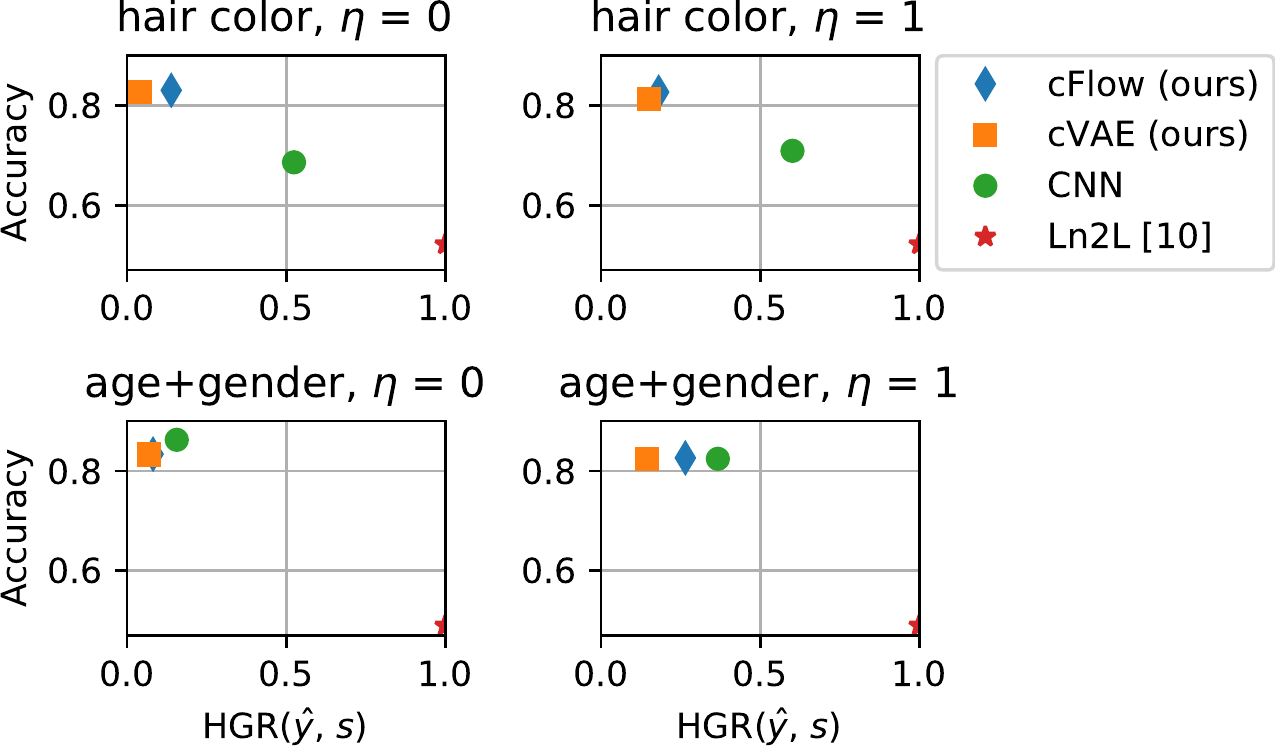}
    \caption{
        For \emph{hair color}, $s$ takes on the values Blond, Brown and Black.
        For \emph{age+gender}, $s$ takes on the values Young/Female, Young/Male, Old/Female and Old/Male.
    }%
    \label{fig:multi-s}
\end{figure}
\noindent In addition to binary sensitive attribute $s$,
we also investigate multinomial $s$ in the CelebA dataset.
First, we do experiments with hair color, where $s$ has three possible values:
blond hair, brown hair and black hair.
The other experiment is with a combination of age and gender,
where $s$ has four possible values, each of which is a combination of a gender and an age:
Young/Female, Young/Male, Old/Female and Old/Male.
To evaluate the fairness for multinomial $s$, we use the Hirschfeld-Gebelein-R\'enyi Maximum Correlation Coefficient (HGR) \cite{mary2019fairness} that is defined on the domain $[0, 1]$ and gives $\text{HGR}(Y,S)=0$ iff $Y \perp S$
and 1 if there is a deterministic function to map between them.
Results can be found in Figure~\ref{fig:multi-s}.

\subsection{Investigation into the size of $z_b$}\label{ssec:size-of-zb}
\begin{table}[ht]
\caption{Results on the CelebA dataset with different sizes of $z_b$.}
    \label{tab:zs-ablation}
    \centering
\begin{tabular*}{\textwidth}{l@{\extracolsep{\fill}}lrr}
\toprule
 $|z_b|$ & $|z_b|/|z|$ &  Accuracy &   DP diff \\
\midrule
          1 &             0.0082\% &  0.60 &  0.63 \\
          3 &             0.0245\% &  0.60 &  0.63 \\
          5 &             0.0410\% &  0.84 &  0.12 \\
         10 &             0.0820\% &  0.84 &  0.12 \\
         30 &             0.2442\% &  0.74 &  0.23 \\
         50 &             0.4070\% &  0.68 &  0.27 \\
\bottomrule
\end{tabular*}
\end{table}
\noindent In the cFlow model, the size of $z_b$ is an important hyperparameter which can affect the result significantly.
Here we investigate the sensitivity of the model to the choice of $z_b$ size.
Table~\ref{tab:zs-ablation} shows accuracy and fairness (as measured by \emph{DP diff}) for different sizes of $z_b$.
The results show that both too large and too small $z_b$ is detrimental.
However, they also show that the model is not overly sensitive to this parameter:
both sizes 5 and 10 achieve nearly identical results.

\subsection{Additional fairness metrics}\label{ssec:additional-fair-metr}
\begin{table}[]
    \caption{
        Additional fairness metrics for the experiments on the CelebA dataset (Fig. 5 from the main text).
        \emph{TPR diff.} refers to the difference in true positive rate.
        \emph{TNR diff.} refers to the difference in true negative rate.
        \textbf{Left:} $\eta = 0$. \textbf{Right:} $\eta=1$.
    }
    \label{tab:my_label}
    \resizebox{.49\textwidth}{!}{
    \begin{tabular}{lrrrr}
\toprule
     Method &  Accuracy &  DP diff &  TPR diff &  TNR diff \\
\midrule
      cFlow &      0.83 &     0.10 &      0.15 &      0.25 \\
       cVAE &      0.82 &     0.05 &      0.09 &      0.18 \\
        CNN &      0.61 &     0.63 &      0.70 &      0.64 \\
 Ln2L\cite{ln2l} &      0.52 &     0.00 &      0.00 &      0.00 \\
\bottomrule
\end{tabular}}
\hfill
\resizebox{.49\textwidth}{!}{
\begin{tabular}{lrrrr}
\toprule
     Method &  Accuracy &  DP diff &  TPR diff &  TNR diff \\
\midrule
      cFlow &      0.82 &     0.33 &      0.28 &      0.21 \\
       cVAE &      0.81 &     0.16 &      0.10 &      0.05 \\
        CNN &      0.67 &     0.75 &      0.66 &      0.76 \\
Ln2L\cite{ln2l} &      0.51 &     0.08 &      0.06 &      0.09 \\
\bottomrule
\end{tabular}}
\end{table}
\noindent In addition to \emph{DP diff}, we report here the result from other fairness measures.
These results are from the same setup as those reported in the main paper.
We report the difference in true positive rates (TPR) between the two groups (male and female), which corresponds to a measure of Equality of Opportunity,
and the difference in true negative rates (TNR) between the two groups.

\section{Optimisation Details}\label{sec:optimisation-details}
\noindent All our models were trained using the RAdam optimiser \cite{liu2019variance} with learning rates $3\times10^{-4}$ and $1\times10^{-3}$ for the encoder/discriminator pair and classifier respectively. A batch size of 128 was used for all experiments.

We now detail the optimisation settings, including the choice of adversary, specific to each dataset. Details of the cVAE and cFlow architectures can be found in Table \ref{tab:vae_architectures} and Table \ref{tab:inn_architectures}, respectively.

\subsection{UCI Adult} 
For this dataset our experiment benefited from using null-samples as inputs to the adversary of the cFlow model. Unlike for the image datasets, we found a single adversary to be sufficient. This was realised as a multi-layer perceptron (MLP) with one hidden layer, 256 units wide. The INN performs a bijection of the form $f: \mathbb{R}^n \rightarrow \mathbb{R}^n$. However, the adult dataset is composed of mostly discrete (binary/categorical) features. To achieve good performance, we found it necessary to first pre-process the inputs with a pretrained autoencoder, using its encodings as the input to the cFlow model, as well as to the adversary. The learned representations were evaluated with a logistic regression model from scikit-learn \cite{scikit-learn}, using the standard settings. All baseline models were trained for 200 epochs.
The Ln2L \cite{ln2l} and MLP baselines share the architecture of the cVAE's encoder, only with a classification layer affixed.

\subsection{Coloured MNIST}
Each level of the architecture used for the downstream classifier and na\"ive baseline alike consists of two convolutional layers, each with kernel size 3 and followed by Batch Norm \cite{ioffe2015batch} and ReLU activation. For the Ln2L baseline, we use an a setup identical to that described in \cite{ln2l}. Each level has twice the number of filters in its convolutional layer and half the spatial input dimensions as the last. The original input is downsampled to the point of the output being reduced to a vector, to which a fully-connected classification layer is applied.

To allow for an additional level in the INN (the downsampling operations requiring the number of spatial dimensions to be even), the data was zero-padded to a size of $32\times32$. The cVAE and cFlow models were trained for 50 and 200 epochs respectively, using $\ell_2$ reconstruction loss for the former. The downstream classifier and all baselines were trained for 40 epochs. For both of our models, an ensemble of 5 adversaries was applied to the encodings, with each member taking the form of a fully-connected ResNet, 2 blocks in depth, with SeLU activation \cite{klambauer2017self}. The adversaries were reinitialised independently with probability $0.2$ at the end of each epoch. While the adversaries could equally well take  null-samples as input, as done for the Adult dataset, doing so requires the performing of both forward and inverse passes each iteration, which, for the convolutional INNs of the depths we require for the image datasets, introduces a large computational overhead, while also showing to be the less stable of the two approaches in our preliminary experiments.

\subsection{CelebA} 
The downstream classifier and na\"ive baseline take the same form as described above for cMNIST, but with an additional level with 32 filters in each of its convolutions at the top of the network. For this dataset we adapt the Ln2L model by simply considering it as an augmentation the na\"ive baseline's objective function, with the entropy loss applied to the output of the final convolutional layer. These models were again trained for 40 epochs, which we found to be sufficient for convergence for the tasks in question. The cVAE and cFlow models were respectively trained for 100 epochs and 30 epochs, using $\ell_1$ reconstruction loss for the former. Compared with cMNIST, the size of the adversarial ensemble was increased to 10, the reinitialisation probability to 0.33, but no changes were made to the architectures of its members.

\subsection{The Pitfalls of Adversarial Training}\label{ssec:pitfalls}
Adversarial learning has become one of the go-to methods for enforcing invariance in fair representation learning \cite{ganin2016domain} with MMD \cite{LouSweLi15} and HSIC \cite{QuaShaTho19}, being popular non-parametric alternatives.
\cite{ganin2016domain} proposed adversarial learning for domain adaptation problems, with \cite{edwardsstorkey} soon after making this and learning a representation promoting demographic parity.
The adversarial approach carries the benefits of being both efficient and scalable to multi-class categorical variables, which many sensitive attributes are in practice, whereas the non-parametric methods only permit pair-wise comparison.

However, when realised as a neural network, the adversary is both sensitive to the values of the inputs as well as their ordering (though exchangeable architectures, such as \cite{zaheer2017deep} do exist, but which sacrifice expressiveness).
Thus, it can happen that the representation learner optimises for the surrogate objective of eluding the adversary rather than the real objective of expelling $s$-related information.
Moreover, the non-stationarity of the dynamics can lead to cyclic-equilibria, irrespective of the capacity of the adversary.

When working with a partitioned latent space, this behaviour can be averted by instead encouraging $z_b$ to be predictive of $s$, acting as a kind of information ``sink``, as in \cite{JacSmeOya18}.
However, this does not have the guarantee of making $z_u$ invariant to $s$ - there are often many indicators for $s$, not all of which are needed to predict the label perfectly.
Training the network to convergence before taking each gradient step with the representation learner is one way one to attempt to tame the unstable minimax dynamics \cite{feng2019learning}.
However, this does not prevent the emergence of the aforementioned cyclicity.

We try to mitigate the aforementioned degeneracies by maintaining a diverse set of adversaries, as has shown to be effective for GAN training \cite{durugkar2016generative}, and by decorrelating the individual trajectories by intermittently re-initialising them with some small probability following each iteration.

\subsection{Tuning the Partition Sizes.}\label{ssec:partition-size-matters}
There are several ways of ensuring that $z_b$ is of sufficient size to capture all $s$ dependencies, while at time is minimised such as that information unrelated to $s$ is maximally preserved. We adopt the straightforward search strategy of, starting from some initial guess, calibrating the value according to accuracy attained by a classifier trained to predict $s$ from $z_b$ on a held-out subset of the representative set, which is measured whenever the adversarial loss plateaus. If the accuracy is above chance level then that suggests the size of the $z_b$ partition, $|z_b|$, needs to be increased to accommodate more information about $s$. If the accuracy is found to be at chance level then are two possibilities: 1) $|z_b|$ is already optimal; 2) $|z_b|$ is large enough that it fully contains both information $s$ as well as that of a portion of $y$. If the former is true, then perturbations around the current value allow us to confirm this; if the latter is true then decreasing the value was indeed the correct decision.

\section{Synthesising Coloured MNIST}\label{sec:color-details}
\noindent We use a colourised version of MNIST as a controlled setting investigate learning from biased data in the image domain. In the biased training set, each digit is assigned a unique mean RGB value parameterising the multivariate Gaussian from which its colour is drawn. These values were chosen to be maximally dispersed across the 8-bit colour spectrum and are listed in Table \ref{tab:cmnist_rgb_values}. By adjusting the standard deviation, $\sigma$, of the Gaussians, we adjust the degree of bias in the dataset. When $\sigma=0$, there is a perfect and noiseless correspondence between colour and digit class which a classifier can exploit. The classifier can favour the learning of the low-level spurious feature over those higher level features constituent of the digit's class. As the standard deviation increases, the sampled RGB values are permitted to drift further from the mean, leading to overlap between the samples of the colour distributions and reducing their reliability as indicators of the digit class. In the test and representative sets alike, however, the colour of each sample is sampled from one of the 10 distributions randomly, such that colour can no longer be leveraged as a shortcut to predicting the digit's class.

\section{Stabilising the Coupling layers}\label{sec:those-darn-coupling-layers}
\noindent Heuristically, we found that  applying an additional nonlinear function to the scale coefficient of the form
$$
s = \sigma (f(u)) + 0.5
$$
greatly improved the stability of the affine coupling layers. Here, $\sigma$ is the logistic function, which we shift to be centred on 1 so that zero-initialising $f$ results in the coupling layers initially performing an identity-mapping.

\begin{table}[ht]
\caption{Mean RGB values (in practice normalised to $[0, 1]$) parameterising the Multivariate Gaussian distributions from which each digit's colour is sampled in the biased (training) dataset. In the representative and test sets,  the colour of each digit is sampled from one of the specified Gaussian distributions at random.}
\label{tab:cmnist_rgb_values}
\centering
\begin{tabular*}{\textwidth}{l@{\extracolsep{\fill}}lll}
\toprule
Digit & Colour Name & Mean RGB      \\ \midrule
0     & Cyan        & (0, 255, 255) \\
1     & Blue        & (0, 0, 255)   \\
2     & Magenta     & (255, 0, 255) \\
3     & Green       & (0, 128, 0)   \\
4     & Lime        & (0, 255, 0)   \\
5     & Maroon      & (128, 0, 0)   \\
6     & Navy        & (0, 0, 128)   \\
7     & Purple      & (128, 0, 128) \\
8     & Red         & (255, 0, 0)   \\
9     & Yellow      & (255, 255, 0) \\ \bottomrule
\end{tabular*}
\end{table}

\section{Qualitative Results for CelebA}\label{sec:qual-results-celeba}
\noindent Learning a representation alongside its inverse mapping, be it approximate or exact, enables us to probe the behaviour of the model that produced it,
and any biases it may have implicitly captured due to entanglement between the sensitive attribute and other attributes present in the data.
We highlight a few examples of such biases manifesting in the cFlow model's CelebA null-samples in Fig.~\ref{fig:celeba_cflow_suppmat}. In these cases, makeup and hair style have been inadvertently modified during the null-sampling due to the tight correlation between these two attributes and the sensitive attribute, gender, to which we had aimed to make our representations invariant. Additionally, in all highlighted images, the skin tone has changed: from male to gender-neutral, the skin becomes lighter and from female to gender-neutral, the skin becomes darker; in the change from male to gender-neutral, glasses are also often removed.
As the model cannot know that the label is meant to only refer to gender, and not to these other (correlated) attributes,
the links cannot be disentangled by the model.
However, the advantage of our method is that we can at least identify such biases due to the interpretability that comes with the representations being in the data domain.

\begin{figure*}[tb]
  \centering
  \subfloat[Original images.]{%
      \scalebox{0.3}{\includegraphics[width=\textwidth]{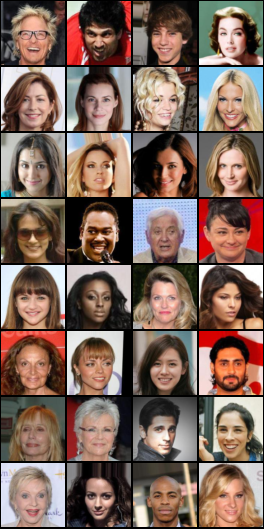}}%
      \label{fig:cvae_celeba_original_x}
  }
  \hfill
  \subfloat[$\bm{x}_u$ null-samples generated by the cVAE model.]{%
      \scalebox{0.3}{\includegraphics[width=\textwidth]{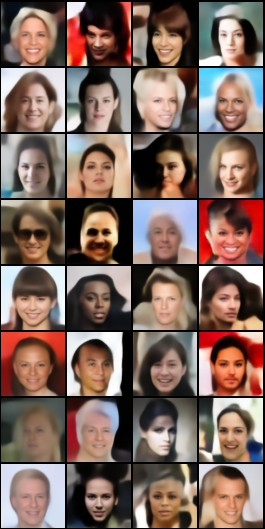}}%
      \label{fig:cvae_celeba_recon_y}
  }
  \hfill
  \subfloat[$\bm{x}_b$ null-samples generated by the cVAE model.]{%
      \scalebox{0.3}{\includegraphics[width=\textwidth]{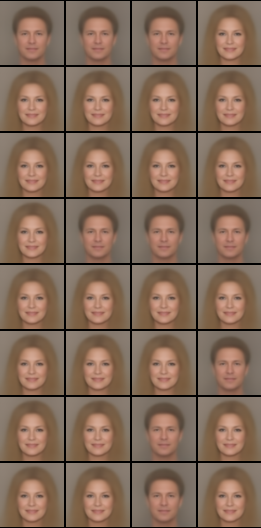}}%
      \label{fig:cvae_celeba_recon_s}
  }
  \caption{
    CelebA null-samples learned by our cVAE model, with gender as the sensitive attribute.
    (a) The original, untransformed samples from the CelebA dataset
    (b) Reconstructions using only information unrelated to $s$.
    (c) Reconstruction using only information related to $\neg s$.
    The model learns to disentangle gender from the non-gender related information. Compared with the cFlow model, there is a severe degradation in reconstruction quality due to the model trying to simultaneously satisfy conflicting objectives.
  }\label{fig:celeba_vae}
\end{figure*}

\begin{figure*}[tb]
  \centering
  \subfloat[Original images.]{%
      \scalebox{0.3}{\includegraphics[width=\textwidth]{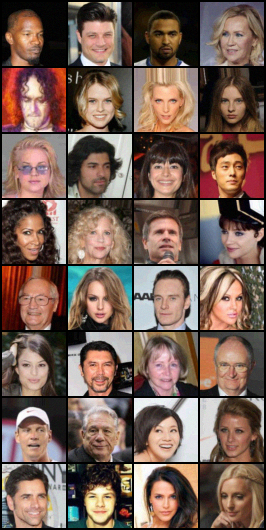}}%
      \label{fig:cflow_celeba_original_x_suppmat}
  }
  \hfill
  \subfloat[$\bm{x}_u$ null-samples generated by the cFlow model.]{%
      \scalebox{0.3}{\includegraphics[width=\textwidth]{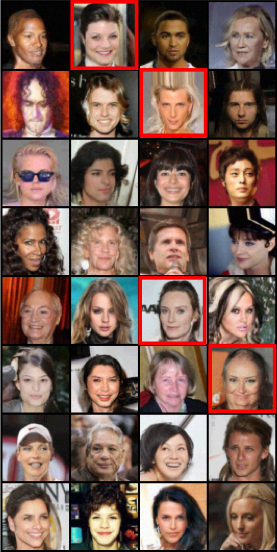}}%
      \label{fig:cflow_celeba_recon_y_suppmt}
  }
  \hfill
  \subfloat[$\mathbf{x}_b$ null-samples generated by the cFlow model.]{%
      \scalebox{0.3}{\includegraphics[width=\textwidth]{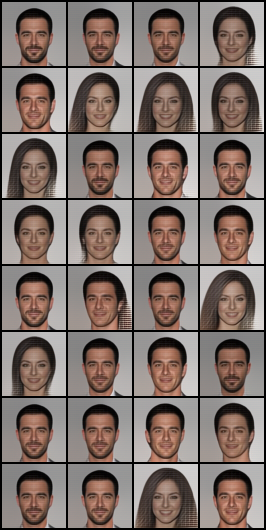}}%
      \label{fig:cflow_celeba_recon_s_suppmat}
  }
  \caption{
    CelebA null-samples learned by our cFlow model, with gender as the sensitive attribute.
    (a) The original, untransformed samples from the CelebA dataset
    (b) Reconstructions using only information unrelated to $s$.
    (c) Reconstruction using only information related to $\neg s$.
    The model learns to disentangle gender from the non-gender related information.
    Attributes such as \emph{makeup} and \emph{hair length} are also often modified in the process (prime examples framed with red) due to inherent correlations between them and the sensitive attribute, which the interpretability of our representations allows us to easily identify.
  }\label{fig:celeba_cflow_suppmat}
\end{figure*}

\section{Transfer Learning}\label{sec:transfer-learning}
For our method, we require a representative set which follows the same distribution as that observed during deployment.
Such a representative set might not always be available.
In such a scenario, we can resort to using a set that is merely \emph{similar} to that in the deployment setting and leverage transfer learning.

One of the advantages of using an invertible architecture over conventional, \emph{surjective} ones that we stressed in the main text is its \emph{losslessness}. Since the transformations are necessarily bijective, the information contained in the input can never be destroyed, only redistributed. This makes such models particularly well-suited, in our minds, for transferring learned invariances:
even if the input is unfamiliar, no information should be lost when trying to transform it.
This works as long as only the information about $s$ ends up in the $z_b$ partition.
If $s$ takes a form similar to that which we pre-trained on, and can thus be correctly partitioned in the latent space, by complement we have the information about $\neg s$ stored in the $z_u$ partition, without presupposing similarity to the $\neg s$ observed during pre-training.

\subsubsection{Transferring from mixed-NIST to MNIST.}
We test our hypothesis by comparing the performance of the cFlow and cVAE models pre-trained on a mixture of datasets belonging to the NIST family, colourised in the same way as cMNIST, while the downstream train and test sets remain the same as in the original cMNIST experiments. Specifically, we create this representative set by sampling 24,000 images (to match the cardinality of the original representative set) from EMNIST (letters only)~\cite{cohen2017emnist}, Fashion\-MNIST~\cite{xiao2017fashion} and KMNIST~\cite{clanuwat2018deep}, in equal proportion. We use the same architectures for the cVAE and cFlow models as we did in the non-transfer learning setting. In terms of hyperparameters, the only change made was to the KL-divergence's pre-factor, finding it necessary to increase it to $1$ to guarantee stability.

\begin{figure*}[htb]
  \centering
  \subfloat[Performance on cMNIST test data after pre-training on the mixed NIST dataset.]{
      \scalebox{0.6}{\includegraphics[width=\textwidth]{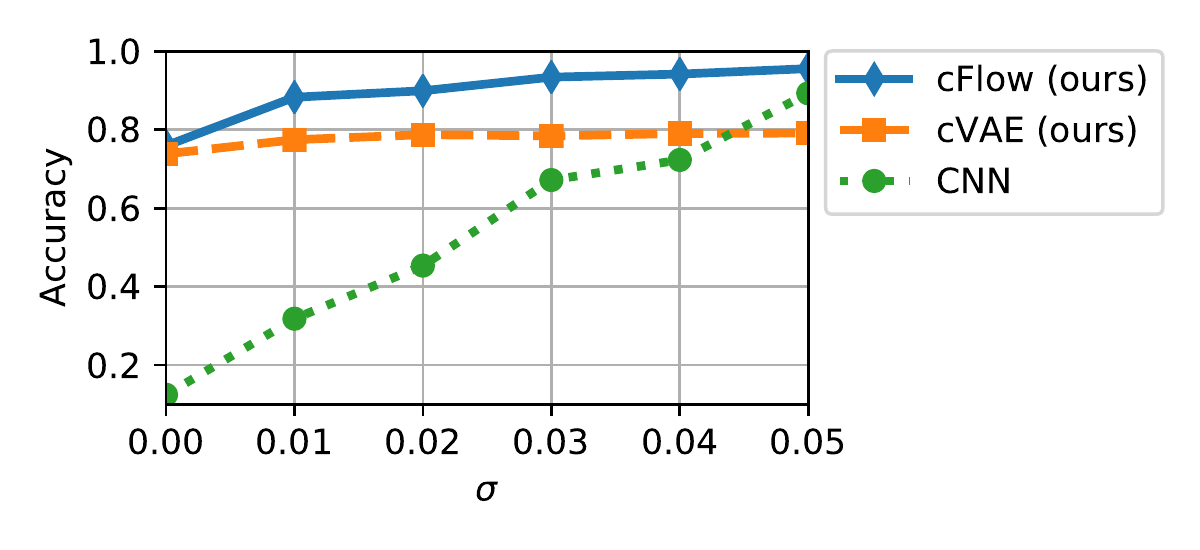}} \label{fig:cmnist-transfer}
  }
  \vspace{10pt}
 
  \subfloat[Test data input to the cFlow model.]{%
      \scalebox{0.3}{\includegraphics[width=\textwidth]{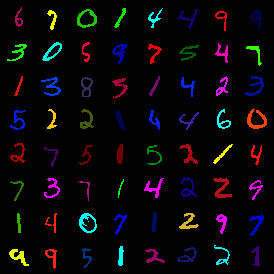}}%
      \label{fig:cflow_tl_original}
  }
  ~~~
  \subfloat[$\bm{x}_u$ null-samples generated by the cFlow model.]{%
      \scalebox{0.3}{\includegraphics[width=\textwidth]{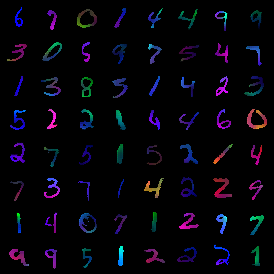}}%
      \label{fig:cflow_tl_xd}
  }
  
    \subfloat[Test data input to the cVAE model.]{%
      \scalebox{0.3}{\includegraphics[width=\textwidth]{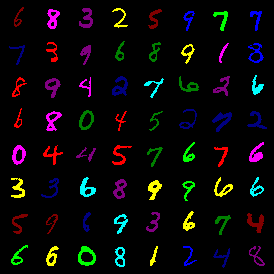}}%
      \label{fig:cvae_tl_original}
  }
  ~~~
  \subfloat[$\bm{x}_u$ null-samples generated by the cVAE model.]{%
      \scalebox{0.3}{\includegraphics[width=\textwidth]{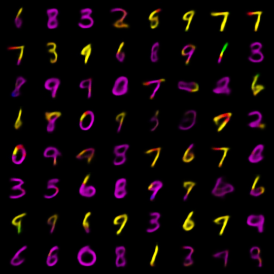}}%
      \label{fig:cvae_tl_xd}
  }
  \caption{
    Results for the transfer learning experiments in which the representative set consists of colourised samples from EMNIST, KMNIST, and FashionMNIST, while the downstream dataset remains as cMNIST. (a) Quantitative results for different $\sigma$-values. (b-c) Qualitative results for the cFlow model.
    (d-e) Qualitative results for the cVAE model. The qualitative results provide comparisons of the images before (left) and after (right) null-sampling. Note that for some of the cVAE samples, the clarity of the digits has clearly changed due to null-sampling, serving as an explanation for the non-increasing downstream performance.
  }%
  \label{fig:cmnist-transfer-all}
  
\end{figure*}

The results for the range of $\sigma$ values are shown in Fig.~\ref{fig:cmnist-transfer}. Unsurprisingly, the performance of both models suffers when the representative and test sets do not completely correspond. However, the cFlow model consistently outperforms the cVAE model, with the gap increasing as the bias decreases.
Although some colour information is retained in the cFlow null-samples, symptomatic of an imperfect transfer, semantic information is almost entirely retained as well.
Conversely, the cVAE is very much flawed in this respect; as can be seen in the bottom row of Fig.~\ref{fig:cmnist-transfer}, for some samples, semantic information is degraded to the point of the digit's identity being altered. As a result of this semantic degradation, the performance of the downstream classifier is curtailed by the noisiness of the digit's identity and is relatively unchanging across $\sigma$-values, in contrast to the monotonic improvement of that achieved on the cFlow null-samples.

\end{appendix}


\FloatBarrier%
\clearpage

\printbibliography%
\end{refsection}

\end{document}